\documentclass[a4paper]{article}
\usepackage[utf8]{inputenc}
\usepackage{graphicx}
\usepackage{titlesec}
\usepackage{caption}
\usepackage{subcaption}
\usepackage{float}
\usepackage{hyperref}
\usepackage{geometry}
\usepackage{amsmath}
\usepackage{csquotes}
\usepackage{amssymb}
\newgeometry{vmargin={25mm,50mm}, hmargin={35mm,35mm}}
\titleformat{\paragraph}
{\normalfont\normalsize\bfseries}{\theparagraph}{1em}{}
\titlespacing*{\paragraph}
{0pt}{3.25ex plus 1ex minus .2ex}{1.5ex plus .2ex}

\usepackage{fancyhdr}

\fancypagestyle{plain}{
\fancyhf{}
\pagestyle{fancy}
\fancyhead[C]{\includegraphics[height=2cm]{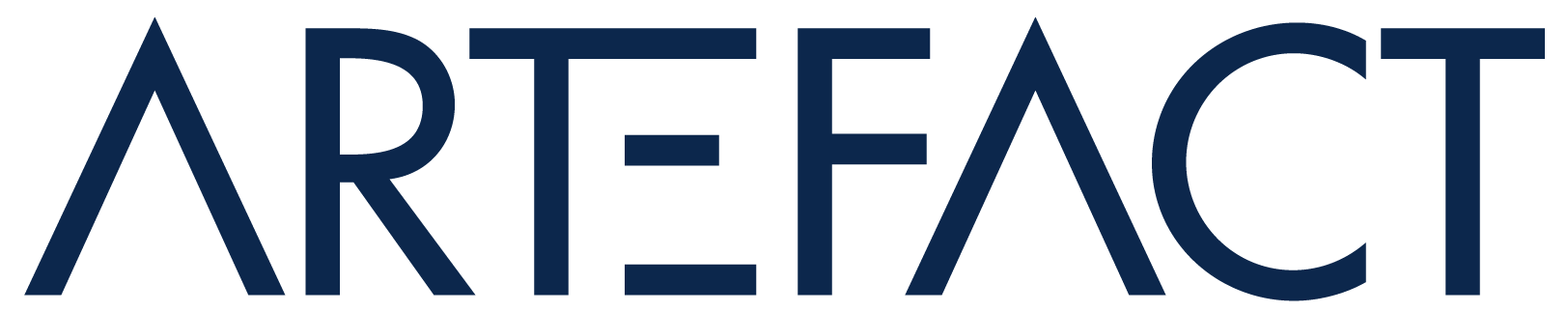}}
\fancyfoot[L]{Copyright \textcopyright \vspace{6pt} 2023 Artefact}
}
\usepackage{amsthm}
\newtheorem{definition}{Definition}
\fancypagestyle{AB}{
\fancyhf{}
\pagestyle{fancy}
\setlength{\headheight}{30pt}
\fancyhead[L]{\includegraphics[height=1.5cm]{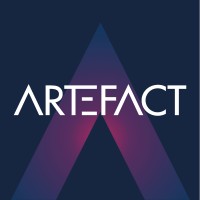}}
\fancyfoot[L]{Copyright \textcopyright \vspace{6pt} 2023 Artefact}
\fancyfoot[R]{ \thepage}
}
\usepackage{authblk}
\title{Instability of computer vision models is a necessary result of the task itself}
\author{  
  Turnbull, Oliver\\
  \texttt{oliver.turnbull@artefact.com}
  \and
  \v{C}evora, George\\
  \texttt{george.cevora@artefact.com}
  }
\affil{Artefact Ltd.\\4th Floor, 1 Lloyd's Avenue London, EC3N 3DS}
\date{}
\begin{document}

\maketitle


\abstract{Adversarial examples resulting from instability of current computer vision models are an extremely important topic  due to their potential to compromise any application. In this paper we demonstrate that instability is inevitable due to a) symmetries (translational invariance) of the data, b) the categorical nature of the classification task, and c) the fundamental discrepancy of classifying images as objects themselves. The issue is further exacerbated by non-exhaustive labelling of the training data. Therefore we conclude that instability is a necessary result of how the problem of computer vision is currently formulated. While the problem cannot be eliminated, through the analysis of the causes, we have arrived at ways how it can be partially alleviated. These include i) increasing the resolution of images, ii) providing contextual information for the image, iii) exhaustive labelling of training data, and iv) preventing attackers from frequent access to the computer vision system.}

\vspace{50pt}

\begin{figure}[H]
    \centering
    \begin{subfigure}[t]{\textwidth}
        \includegraphics[width=\textwidth]{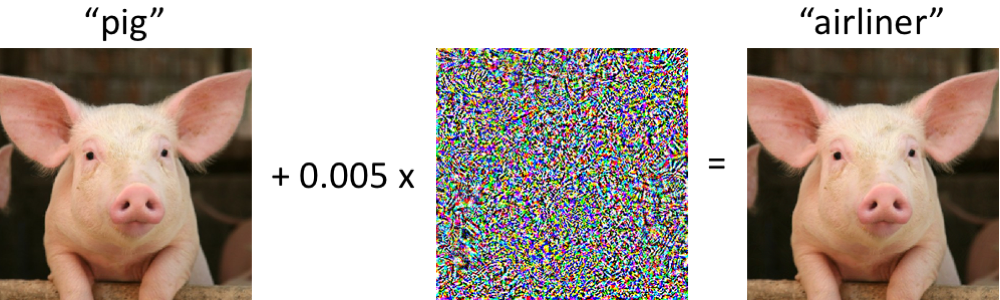}
        \caption{The original image of a pig gets misclassified by a machine vision system as an airliner after an addition of faint (0.005x) but carefully crafted noise-like overlay. \cite{pigliner}}
    \label{fig:pig}
    \end{subfigure}
    \begin{subfigure}[t]{\textwidth}
    \includegraphics[width=\textwidth]{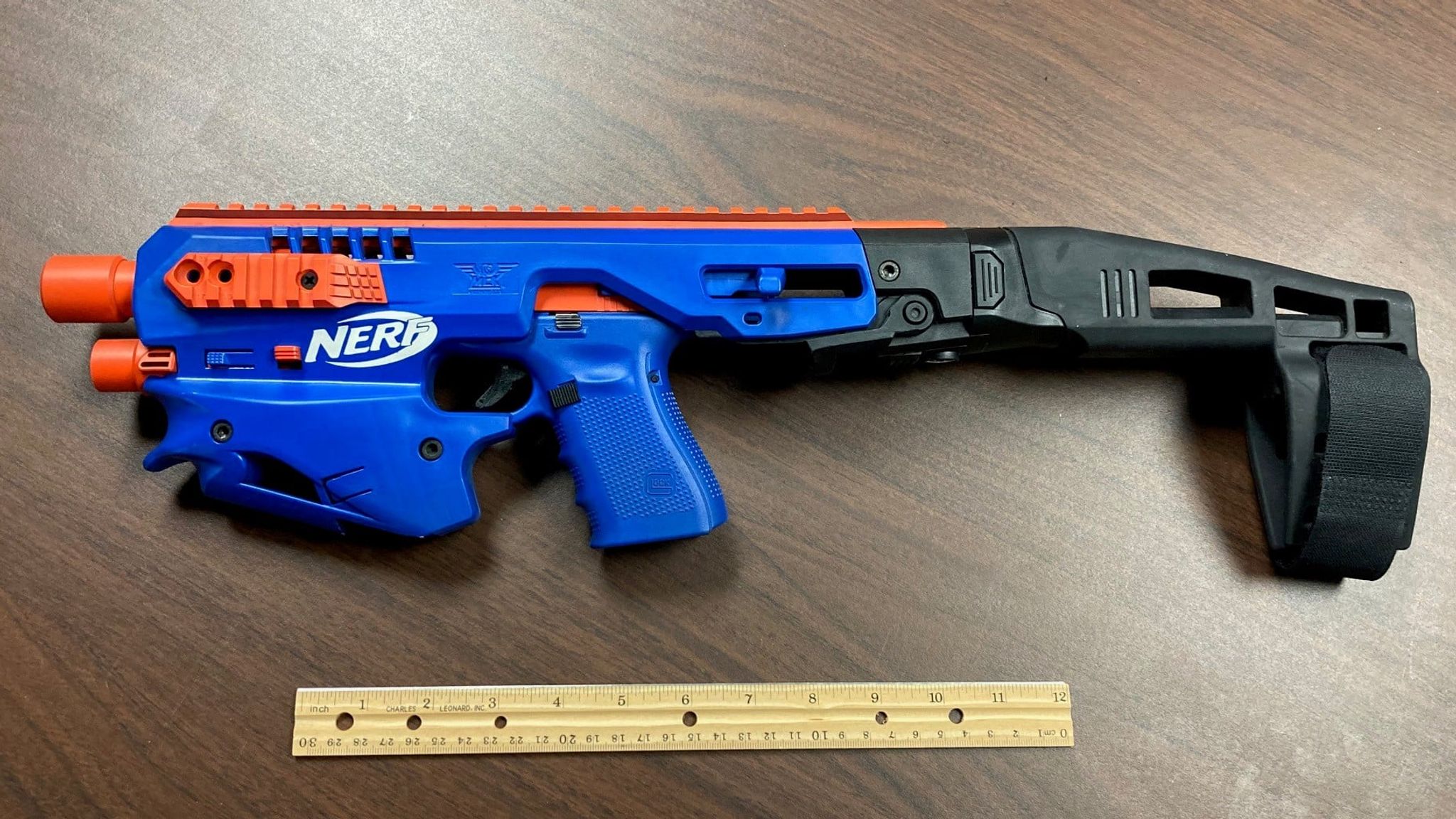}
    \caption{An actual assault rifle disguised as a toy. \cite{NerfGun}}
    \label{fig:nerf}
    \end{subfigure}
    \caption{Adversarial Examples in machine vision are images that the system misclassifies. Some of the examples such as \ref{fig:nerf} seem as reasonable errors to humans, while others such as \ref{fig:pig} are treated as indicative of inferiority of machine vision systems to human vision.}
    \label{fig:adversarialExamples}
\end{figure}

In recent years, there has been growing concern of the impact of adversarial examples on the reliability of computer vision systems, particularly those using deep learning as their core inference engine. These adversarial examples are small changes to an input image that results in a change of classification label from the model. This can look like adding a small amount of noise to an image of a pig and having it mislabelled as an airliner, or paint on the outside of a gun to make it unrecognisable to any system looking to  detect dangerous weapons as shown in Figure \ref{fig:adversarialExamples}. Many \cite{adv_attack_1,Xu2020,he2018decision} researchers have published work on algorithms to generate white box or black box with querying adversarial examples, which begs the question: are computer vision algorithms doomed to fail? Some researchers \cite{goodfellow2014explaining} have identified grounds for the existence of adversarial examples among the features of deep neural networks. In this paper we aim to make a much broader statement: that the existence of adversarial examples is a necessary result of how computer vision problems are formulated.

Evolution has been the source of early examples (chameleon, octopus, praying mantis, etc), with humans later inventing camouflage. We argue that the Adversarial Examples in the Deep Learning context are fundamentally the same concept; as such the DL Adversarial Examples should not be seen as a problem, but as a necessary result of perceptual systems optimised to differentiate between distinct predetermined categories. It is the emphasis on accuracy within the ML framework that is to blame, though, at the same time, we do actually want accuracy. 

Here we may want to distinguish between formal and computational instability for our classification function $f$ and input space $\Omega$. Formal instability being that for any $\epsilon > 0$, we can find an $x \in \Omega$ such that there exists $y \in B_{\epsilon}(x)$ where $f(x) \neq f(y)$, and we can prove that these points exist for \textit{any} finite label classification problem. Computational instability is thus therefore making a similar claim, but placing some other conditions on the $\epsilon$ chosen i.e. that it is larger than some minimum resolution so that images only change label for some ``perceptible" change in input. In this latter case we can imagine that given a sufficiently low measure boundary that given our change sizes a change in output label is still acceptable to the classification problem. Neither of these are necessarily an issue for any classification problem provided that the region of ``instability" is small; small in relation to measure. We would like it to be the case that it is unlikely that an input picked at random would be a formally or computationally unstable point.

\section{The Computer Vision View}

In this section we seek to establish a few key points. Firstly, that fractal levels of instability can occur with even very simple convolution operations. Secondly, that the manifold structure of our input space has fewer samples closer to the boundary and so will necessarily fail at reproducing said boundary. Thirdly, that nondifferentiable activation functions are partly responsible for accelerating this problem, and fourthly that in spite of its necessary existence, instability in a computer vision system is in some sense a non-fundamental issue; i.e. as image resolution increases the effect becomes less and less. 

There are some specific considerations with regards to the structure of computer vision systems. In particular, we take the view that computer vision systems are essentially feature extractors for the input image, and these features represent the semantic content of the image, which is then used as the actual space where the categorical decision is made. This can be done as simply dimensionality reduction on the raw image pixel values, or something like a convolutional neural network where geometrically local dimensions of input feature space are used to constrain how the features get generated. There are thus three places where instability can arise; firstly in the input space itself, secondly in the manifold-feature representation of the input space, and thirdly in the mapping from the feature space to the output space (usually a classification). The problems with the input space we will come back to later, but in this first section we will talk about how feature extraction and feature mapping- particularly in convolutional neural networks, provide the opportunity for instability to occur.

\subsection{Feature Entanglement}
Here we wish to define the type of instability that we consider to be unique to computer vision and particularly convolutional neural network classifiers, and reason about this from a highly simplified lens. We informally define features of a classifier as entangled if there is little distance from the region of one feature's highest expression to the other. Examples of this might be one feature's highest expression having points dense in another region, or pseudodense 
to some tolerance $\epsilon$.
It has been be shown \cite{ilyas2019adversarial} that any kind of feature extraction on images can identify `fragile' features; features that correlate with output label at a point but not everywhere in a neighbourhood around this point.

\begin{figure}
     \centering
     \begin{subfigure}[t]{0.49\textwidth}
         
             \includegraphics[width=\textwidth]{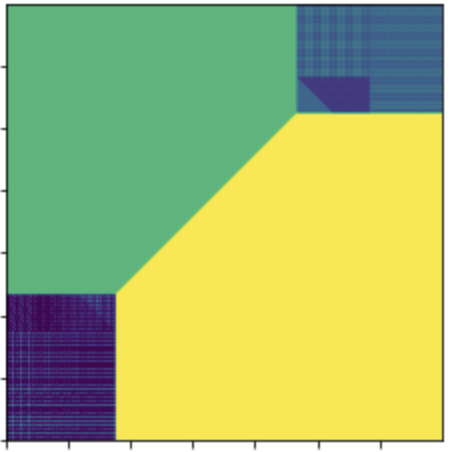}
         \caption{Result of the filtering over unit space.}
         \label{fig:fractal}
     \end{subfigure}
     \hfill
     \begin{subfigure}[t]{0.49\textwidth}
         
         \includegraphics[width=\textwidth]{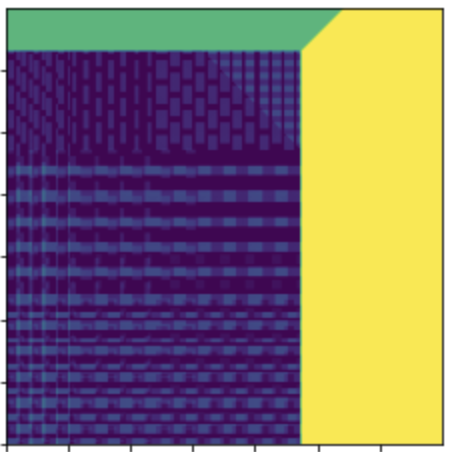}
         \caption{Subsection of the unit space showing clear instability resulting from the filtering.}
         \label{fig:zoomed}
     \end{subfigure}
        \caption{Demonstration of instability resulting from simple filtering. Four vectors used for filtering are $(1/2, 1/2), (2/3, 1/3), (1/3, 2/3)$ and $(1/4, 3/4)$. The vector with lowest L1 norm between itself and a given point in space was selected as a ''winner'' and determine the colour of the point. It is apparent that even such simple filtering produces a great level of instability.}
        \label{fig:three graphs}
\end{figure}

Though, indeed, fragile features can arise from even very simple types of filtering. If instead of a componentwise product followed by a pooling operation (as is the case with the filters in neural networks) we take an L1 similarity measure we can observe something like figures \ref{fig:fractal} and \ref{fig:zoomed}. Here we have taken the similarity to the vectors $(1/2, 1/2), (2/3, 1/3), (1/3, 2/3)$ and $(1/4, 3/4)$ as our filtering, taking the argmax of the dot product of the grid point coordinates with these given tuples. The colour shows which of these is the max value.

In this, we see not just a computational instability but a true unstable region; where points of different highest feature expressions are dense in each other; any neighbourhood of a point here will contain another point of different highest feature expression. 

Maxpooling will create the exact kind of undifferentiability as generated in the L1 filtering, suggesting an inherent tendency to regions of high instability along any one particular feature axis apply to most current CNNs. While the features embedded in a higher dimensional feature space as they are, may not exhibit the same behaviour in conjunction with one another. We will see a similar effect from softmaxing at the final layer, which is a smooth approximation to an L1 max function. This will asymptotically approach the behaviour observed in figure \ref{fig:fractal} as the number of labels predicted and or temperature of the softmax increase.

\subsection{Feature extraction as manifold learning; distribution gradients}

Similarly to Ilyas \cite{ilyas2019adversarial}, we define the classification problem in general as follows
\begin{equation}
     C(x) = \textrm{sign}\bigg(b + \sigma\Big(\sum_{i=0}^{n} w_if_i(x)\Big)\bigg) 
\end{equation}
for some feature map $\vec{f}$. This feature map is a dimensionality reduced representation of the data, with the goal of learning a statistical manifold paramaterised by the components of $\vec{f}$ (in a geometric rather than statistical sense). The final layer, therefore, is an attempt to learn the \textit{statistical} parameters of the distribution at each point; in this case the parameterising probability of a Bernoulli distribution at each $x$.

The idea is thus that these statistical parameters vary continuously across the manifold; with instability being generated by steep gradients in neighbourhoods of certain points in the image of $\vec{f}$. If $C$ is continuous on $\mathbb{R}^n \supseteq \mathcal{M} \rightarrow [0, 1]$, and the classification problem is sharply defined, then we will expect the classifier to have steep gradients at the true boundary. If, as is more likely the case, the problem is not sharply defined (i.e. not every image is a definite yes or no) we may still be finding adversarial examples, depending on the number of labels relative to the dimensionality of input and density of data.

\subsection{Entanglement as a property of Translational Invariance and Feature Equivariance}

There is an important property of computer vision problems and their solution space that could give us a clue as to why, in practice, there are many regions of instability in algorithms that approximate the true boundary. Let us consider the $m \times n \times 3$ grid of pixels that gets processed by machine learning systems; we can think of the problem space as a classification problem on $\mathbb{R}^{3mn}$ that has to preserve labelling under a number of permutations. For a classification space $\mathcal{C} \subset \mathbb{R}^k \times [0,\; 1]$ we call the set (in this particular case actually a group) of permutations on the components of elements of $\mathbb{R}^k$ that preserve labelling $\textrm{Inv}(\mathcal{C})$. For standard computer vision problems, this set contains grid translations, reflections and rotations each of which can be parameterised by at most two vectors (in this case grid coordinates) meaning the size of $\textrm{Inv}(\mathcal{C})$ is $O((mn)^2))$.

Let us take some element of the input space $y \in \mathbb{R}^{3mn}$ with our classification function (true or otherwise!) $f$, then we know the set of all images of $y$ under some invariant transformation is the same size as the set of all transformations: $|\{\alpha(y): \alpha \in \textrm{Inv}(\mathcal{C})\}| = |\textrm{Inv}(\mathcal{C})| = O((mn)^2)$, an order parameter for the size of this set that will come in useful later. Suppose $y$ is $\epsilon$-close to our classification boundary i.e. that a ball of size $\epsilon$ will fit entirely within the classification subspace of $y$ or mathematically, $f(B_{\epsilon}(y)) = \{f(y)\}$. Then we must also have $$f(\{\alpha(x) : \alpha \in \textrm{Inv}(\mathcal{C}), \; x \in B_{\epsilon}(y)\}) = \{1\}$$You can think of this as taking a little ball entirely inside the input space and seeing all its locations under each transformation that we know preserves classification, so we end up with a collection of possibly overlapping balls in the input space for which every point inside has the same classification, chosen here to be $1$ WLOG. If we now consider $z \in f^{-1}({0})$ then we know, if we call $\mathcal{X} = \{\alpha(x) : \alpha \in \textrm{Inv}(\mathcal{C}), \; x \in B_{\epsilon}(y)\}$  that $z$ is $\nu$-close to the boundary where $\nu = \textrm{min}(\{|z-x|: x \in \mathcal{X}\})$, with the immediate result that $\nu \leq |z-y|$. What this means is that as the dimension of the input space (in this case resolution of the image) increases, we get more transformations on our input space that preserve labeling and so more points that could be closer to the boundary than a given known classification point; adding to the instability of our problem.

However, if we take a standard result from measure theory \cite{unit_ball}, we know that if $B_\epsilon(y) \subset \mathbb{R}^k$, then $$\textrm{Volume}(B_\epsilon(y)) = \frac{\pi^{k/2}}{\Gamma(\frac{k}{2} + 1)}\epsilon^k$$ So the volume of our set under our group actions becomes $$\textrm{Volume}(\mathcal{X}) \leq |\textrm{Inv}(\mathcal{C})|\frac{\pi^{k/2}}{\Gamma(\frac{k}{2} + 1)}\epsilon^k$$ Immediately we see that the volume of this set tends to zero so long as $|\textrm{Inv}(\mathcal{C})|\frac{\pi^{k/2}}{\Gamma(\frac{k}{2} + 1)}| \rightarrow 0$, and so even though the number of unstable points increases, the probability of falling within $\epsilon$ of one of these points tends to 0. Since we know that for computer vision problems, we have $|\textrm{Inv}(\mathcal{C})| \in O((mn)^2) = O(k^2)$, then we are happy that we are not likely to fall within $\epsilon$ of a given boundary point.

Of course, we do not have just one point $\epsilon$-close to our boundary, but we do have that our total measure will be bounded by integrating over a ball of radius $\epsilon$ for each point on the boundary. But it is important to note that in the context of computer vision, we have maximum and minimum intensities of pixel values, meaning our input space is restricted to (something homeomorphic to) the unit cube and as such the total measure will be bounded by 1, so we can reason directly from our ratio still.  This tells us that while there may be an inherently greater level of instability in the instance of computer vision problems than others, under a high enough resolution this effect becomes small.

This then gives us an interesting extension; while the classes of symmetries are sufficiently small for computer vision problems that this isn't an issue, this becomes an issue for problem spaces with a higher degree of permutation invariance, namely, graph neural networks. In such cases, our bound (given the number of permutations is $k!$) does not tend to 0, which suggests it is possible to cover our entire input space with neighbourhoods of unstable points and their neighbourhoods under transformations.

We can also apply this argument to symmetries that exist within a classifier i.e. that softmax of the feature layer is itself invariant under permutation. The preimages of these feature vectors will in turn be responsible for many $\epsilon$-balls in input space and as such increase the instability of the space. Let us take a feature vector $\vec{f}$ as the output of some classifier layer between the input and output spaces. If the input space is $k$ dimensions and the output space is $r$ dimensions then we can expect each $\vec{f}$ to correspond to on average $\frac{k}{r}$ points in input space. So in the case $\vec{f}$ is $\epsilon$-close to the decision boundary in feature space, it corresponds to both $r$ points in feature space thus and $r\frac{k}{r} = k$ points in input space, giving us another polynomial degree term in our total measure calculation, preserving its tending to $0$.

\subsection{Conclusion}
We have established that while adversarial examples that arise from symmetries of a particular problem or classification algorithm do exist, their proportion in relation to points further away from the boundary shrinks to zero eventually. Moreover, this kind of symmetric analysis can be used to analyse all kinds of neural network architectures, namely where we see that for graph neural networks (which have a much larger space of classification-invariant transformations) that no such limiting process exists and the probability of finding an adversarial example may tend to 1 as the input space gets bigger in dimension. 

Presence of adversarial examples, however, does not present a problem per se. It is the ability of hostile actors to find them that causes practical issues. The current methods  of finding adversarial examples require either access to the machine vision system itself or ability to query it repeatedly. A simple protection from adversarial attacks is therefore denying access to the system itself for potentially hostile actors. If a machine vision system in question is required to be exposed to public, via interface such as an API the solution is to prevent hostile actors from performing  sufficient number of queries to find adversarial examples - for instance by API throttling.

\section{The Categorical View}
Beyond the behaviour of the mathematical and computation models of image recognition, we have the behaviour of classification algorithms more generally. Here we will explore some of the limitations of any kind of categorisation problem on certain classes of spaces, outside of the specifics of the algorithm, problem subspace, or type of data. We will establish that instability in a formal sense will always occur in any continuous classification problem, when this instability becomes a problem in practice, and to what extent we can foresee and interrupt situations where it may be a problem by looking at the structure of the spaces used in these problems. We will finish with a brief philosophical discussion of the relationship between an object and its image, and how this relationship might affect systems built to recognize objects from images.


It can be shown that for any compact topological space (which we discuss in the appendix), a continuous map from it to a discrete set of values means that there will either be an image mapped to more than one label, thus the resulting map is not a function, or that there is a point mapped to no label. Since we are presuming the former never occurs, i.e. our output is repeatable and not stochastic, there must be points that are not mapped to either label; the set of such points we call the boundary of our classification problem. If we add metric structure to the compact topological space, then we know that there will exist non-boundary points close to these boundary points, necessitating the existence of unstable points (again, this can be justified with a short proof). When we talk about stable classifiers in a formal sense (a small change in input corresponds to a small change in output) we mean they are are continuous in a topological sense. 

The presence of adversarial examples does not necessitate instability in the colloquial sense: the presence of a classification boundary necessitates the existence of points epsilon-close to that boundary. Instead for practical reasons we are concerned about far-deviations from emph{calibrated probability} guesses near the boundary - i.e. being certain when the result is certain and uncertain when not.

\subsection{Humans are No Better}
However, beyond the problems with classification in general, we must understand that this problem is situated under comparison with humans. It is not simply that we want to understand classification instability in the abstract, which is its own necessity, but that we want classification to be unstable in exactly the same way as human classification is. While this notion is rarely spelled out in the literature, it seems to underpin the wider discussion of the topic.

\begin{figure}[ht]
    \centering
    \begin{subfigure}[t]{0.49\textwidth}
    \includegraphics[,width=\textwidth]{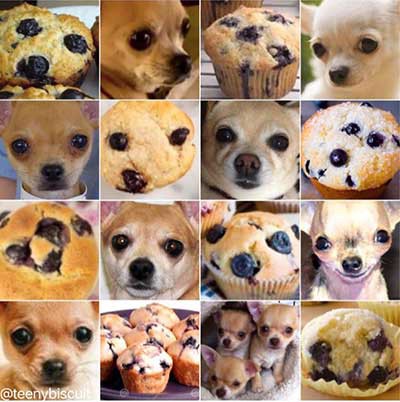}
    \caption{Is it a dog or is it a muffin? \cite{chihuahua}}
    \label{fig:chi_muff}
    \end{subfigure}    
\hfill
    \begin{subfigure}[t]{0.49\textwidth}
    \centering
    \includegraphics[,width=\textwidth]{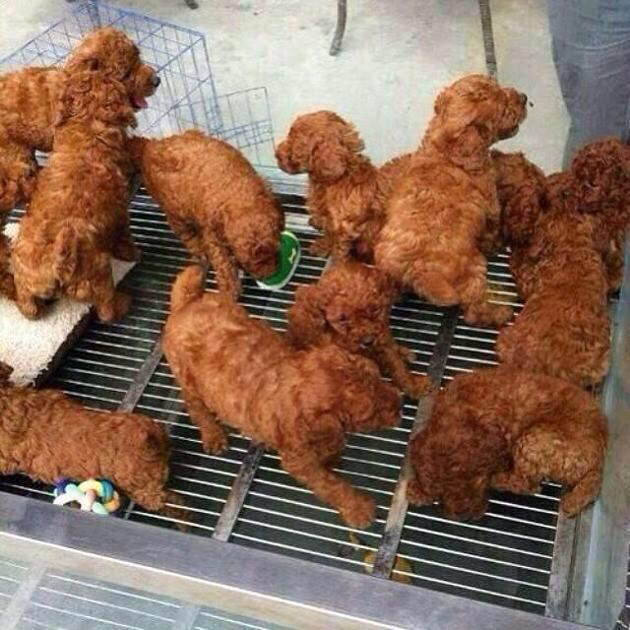}
    \caption{Is it a dog or is it fried chicken? \cite{dogchick}}
    \label{fig:dog_chick}
    \end{subfigure}
    \caption{While the topic of adversarial examples is rarely related to human vision it is useful to realize \emph{visual illusions} can be considered adversarial examples to human vision system.}
    \label{fig:illusions}
\end{figure}

As demonstrated in figure \ref{fig:illusions} humans too have regions of indeterminacy where small changes to the input features change confidence in labeling i.e. our calssification boundary is not measure 0 either. This gives us a kind of hedge against the embarrassment of computer vision systems; human errors in classification are considered a natural feature of our visual system while machine vision errors are considered failures at the task.

Yet, for some reason, we have a sense that humans, in their image classifications, do not have the same problem of instability as we would not change our labelling from an obvious example to the opposite label with a small perturbation. Of course, any classifier (humans included!) will have unstable points. However,  we believe that our unstable points are in some sense better than the examples such as the pig/airliner adversarial example shown in figure \ref{fig:pig}. 
    
    We know that one model's string classification could easily be another model's edge case; indeed that where decision boundaries fall can be very different even when very similar around the training points. However, we also have a sense of intelligibility about the content of an image (nonzero measure boundary).

    Consider the case where two classifiers are competing over an image containing both cat and dog, one classifying the image as a cat, the other one as a dog. Which of these is the correct classification of an image containing both? 
    The problem arises when our algorithms have been trained on single label images only. In order for our labelling to be as accurate as possible we require that we know when images of pigs do not have airliners in them and vice verse so that we can capture these sorts of relationships that might appear. But this is not representative of the datasets used in pracitce. If we treat each image as only containing one thing, then we may be introducing instability where there isn't really any at all. 
    This prompts a nice solution in first specifying your ontology (collection of all possible labels) and training classifiers on the images simultaneously, so that the statistical noise in other classifiers (that may pick up on some significant feature here or there in the background of an image) do not counteract the presence of other objects or features in the image.

\subsection{Ceci n'est pas le sujet}

\begin{figure}[ht]
    \centering
    \includegraphics[scale=0.3]{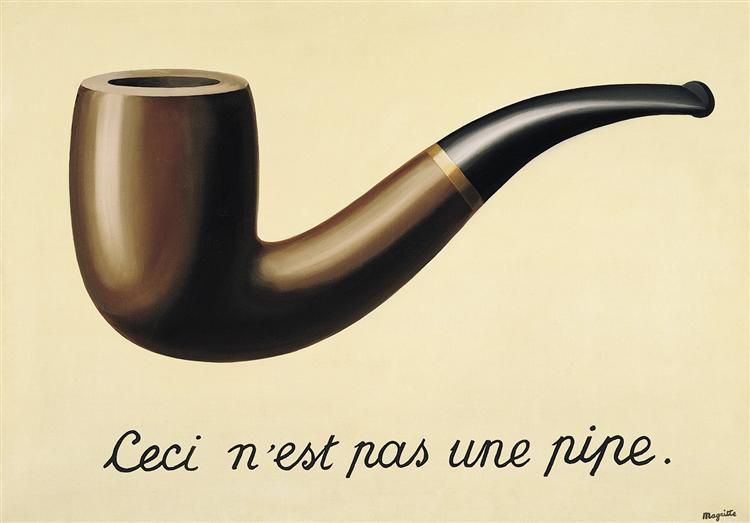}
    \caption{The treachery of images is a 1929 painting by Ren\'{e} Magritte. \cite{treachimage} It is generally understood to point out that the painting is not the painted object itself.}
    \label{fig:pipe}
\end{figure}

One last point to mention, a subtle but important one, is that the image of something is not the thing itself. In our classification problem we are trying to construct a probability distribution for an image coming from a thing we are labeling, but the object of our classification is ultimately a photograph of that thing - concept discussed by Ren\'{e} Magritte in his seminal painting Trechery of the images shown in Figure \ref{fig:pipe}.  Thus we fail to capture the full range of features that might determine a classification and allow for completely different features to be flattened into very similar representations (such as wings and ears being similar in images but very different in real life). The most common example of this is scale; an image is often composed so that the object of it is at the center; meaning that the classification of that object does not take into account the size of the object itself, or the other objects it is surrounded by.

The image presupposes a framing and a context that may create similarity between two actually very different scenes. The image itself is just an image, and so any instability in the classification may come from a fundamental underparamaterisedness of the problem for computer vision; the classifiers do not understand the relationships between the things they are trying to predict; they have no understanding of, to use an earlier example, the relation between pigs and airliners, they simply identify statistically significant features of them each. In a similar sense, the number of possible three-dimensional scenes that could give rise to a given image is potentially infinite; objects occluded by foreground objects could be anything, meaning always vital context that would come from the exploration of the object in the same space humans experience it is lost when a photograph is captured. The image could be of a deceptive actor holding up an image of something else to the camera, even occluding the camera's view altogether, meaning the supposed true nature of the image (that it is of a person holding a piece of paper) is lost entirely to the viewer of the image (both human and algorithm, since there cannot be a distinction between replication of an image and an image representing a real thing - especially digitally, where all images are copies in some sense).

\begin{figure}[ht]
    \centering
    \includegraphics[scale=0.2]{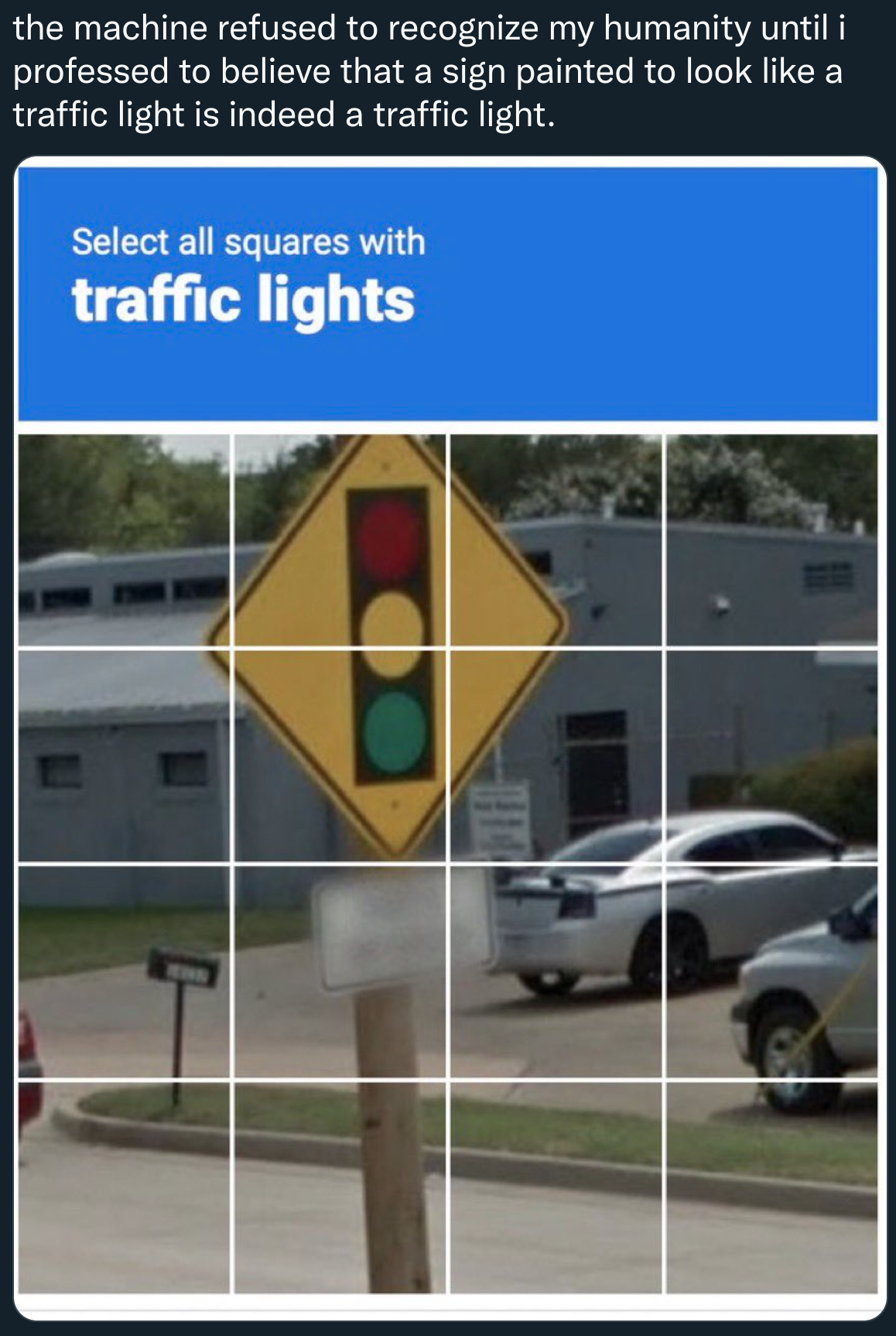}
    \caption{Machine itself is fooled by the image demonstrating the impossibility to distinguish between object and its depiction purely from a single snapshot. \cite{InhumanCapatcha}}
    \label{fig:lights}
\end{figure}

Herein lies what you might call a hubris of prediction; that the most likely thing is always the \emph{object of belief}. This term is a kind of frequentist bastardisation of Occam's razor; that the image represents an actual fixed and determinate thing rather than a probability distribution over all possible things, situations, known unknowns and unknown unknowns. To the contrary, this inherent uncertainty and indeterminacy in the image, and in some sense all problems of prediction, is something we must take into account when deciding action on the basis of prediction. In practice if  systems such as weapon detection computer vision systems aren't treated with the appropriate epistemic humility, then the consequences for both harm reduction as well as the impact on peoples' lives could be severe. Instability exists in computer vision in part at least because vision itself is unstable; what we see and what we interpret from what we see is not always the truth. We must build in contingency for the random failures that always occur in technological systems, both machine learning and traditional if we ever want these systems to be functional in a complex, imperfect, and fluctuating world. As Baudrillard, a philosopher of simulacra and images said in \cite{baudrillard}
\begin{displayquote}
``I would
like to conjure up the perversity of the relation
between the image and its referent, the supposed
real; the virtual and irreversible confusion of the
sphere of images and the sphere of a reality
whose nature we are less and less able to grasp.
There are many modalities of this absorption,
this confusion, this diabolical seduction of
images. Above all, it is the reference principle of
images which must be doubted, this strategy by
means of which they always appear to refer to a
real world, to real objects, and to reproduce
something which is logically and chronologically
anterior to themselves. None of this is true. As
simulacra, images precede the real to the extent
that they invert the causal and logical order of
the real and its reproduction."
\end{displayquote}
In the computer vision system, we presuppose no distinction between real subject and image, meaning that an image of an image as depicted in figure \ref{fig:lights} or \ref{fig:pipe} can never be depicted from simply an image since the image is never made distinct from its subject. 

We must understand the image as a lower dimensional representation of what it represents not only literally (two dimensional rather than three) but also conceptually; the stop sign exists in relation to a road, it has different forms in different countries, it has a material that it is made of and is only placed at specific places on the road. Without this categorical information, the image classification algorithm is already working at a far lower dimensionality and with far less data than any live animal let alone a human, and to truly increase its resolution enough to compete with a human's vision it will need to include these conceptual dimensions as well, not just spatial and colour ones. Not only this but these extra dimensions reduce the symmetries in the problem, meaning destabilisation can occur at lower dimensions; a form of categorical regularisation.


In conclusion we understand instability as arising from a lack of context for an image, from the fact that by the framing of the task, the image of the thing is identical to the thing itself, and that adding context amounts to increasing the dimensionality of the problem and removing possible symmetries from the input space at the same time.

\section{Conclusion}

 The framing of computer vision instability we have established offers some interesting protections against the types of adversarial attacks we see in practice, as well as establishes some a priori limits on the efficacy of these attacks. Specifically protection from adversarial attacks can be increased through:
 \begin{itemize}
 \item Increasing the resolution of images.
 \item Providing contextual information for the image.
\item Exhaustive labelling of training data.
\item Preventing attackers from frequent access to the system.
 \end{itemize}
 
\subsection{Is this behaviour undesirable?}
If we ever want computer vision systems to not only replicate but outperform human classification, we will need them to identify ever more fragile and esoteric features of the data. These are features that are statistically significant at some point in feature-space but fall off quickly in a neighbourhood of that point. Our primary concern is that these don't happen in commonly acceptable regions of input space. We don't have any way of visualising what the true boundary of these classification spaces is, and our ground truth labelling is always provided by humans in some way anyway. Not only this, but computer vision systems must have contingencies for the ways in which images inherently fall short of representing the object of the image, which are precisely where these fragile features will arise. This is, however, lucky for computer vision. Other types of data have a degree of symmetry much higher than an image, such as a graph, for which there is not always a clear way to increase its resolution but further that even if there were this would increase the instability inherent in the problem.

\subsection{Further steps}
In this article, we have shown that in many ways instability is inevitable. We have also shown that there are types of problems that by virtue of symmetry in the problem lend themselves to instability, and propose this as a tool for analysing many other machine learning tasks and algorithms. Further to this, we have tried to understand the inclusion of extra data in these problems as a way of breaking symmetries of the problem, thus making both the problem and associated algorithms more stable. We can do this through understanding that the images of our classifications may cover a range of phenomena (such as the "image-of-image" as described earlier, and that our outputs could be distributional or Bayesian in nature. In the future, we must make sure that our detection or regression algorithms take place at a discriminative enough resolution, and that epistemic uncertainly in build into the output distributions that lay the foundations for our "predictions of truth", predictions always predictaed on risk.

\bibliography{bib}
\bibliographystyle{plain}

\newpage
\section{Appendix}

\begin{definition}{Strict Classifier}
\end{definition}
We say a classifier is \textit{strict} if its output space is discrete. Thus, a classifier is \textit{non-strict} if it outputs a continuous variable (i.e. a probability $p \in [0, \; 1]$)

\begin{definition}{Stability}
\end{definition}
We say a strict classifier $f$ is stable on an input space $\Omega$ if 
\begin{align*}
    \forall x \in \Omega,\; U \in \mathcal{T}(\Omega),\; x \in U: \; \exists V \subset U \; \textrm{s.t.} \; x \in V, \; V \in \mathcal{T}(\Omega) \; \textrm{and} \; f(V) = \{f(x)\}
\end{align*}

i.e. every point has an open neighbourhood that does not change label \break

An immediate is that a strict binary classifier can only be stable on a disconnected topological space, meaning if your input space is connected there must be some points that will change label in a neighbourhood under \textit{any} perturbation. This is by the topological definition of continuity which says:

\begin{definition}{(Continuity):}
    We say a mapping $f:X \rightarrow Y$ is continuous if whenever $U$ is an open set in $Y$ then $f^{-1}(U)$ is open in $X$
\end{definition}

We see that a classifier is stable if it is continuous on $\{0, \; 1\}$ in the topology where $\{0\}$ and $\{1\}$ are open sets, and so you could express $\Omega$ as the union of two nonintersecting open sets: $f^{-1}(\{0\}) \bigcup f^{-1}(\{1\})$. We show these are nonintersecting by taking for all $x \in \Omega$ a $U_x \in \mathcal{T}$ such that $f(U_x) = \{0\}$. Then we see $f^{-1}(\{0\}) = \bigcup_{x \in \Omega}U_x$ since it contains and is contained by it, since we have used every $x$ that gets mapped to $0$ in our construction and no element is mapped to $1$ by the property of it being stable. \break

This is not the case in soft classification. In soft classification we can define stability formally as $f$ being continuous. This is, perhaps, an overly strict definition of stability for the case of computational tractability, where we often have a minimum resolution changes can be seen at meaning that even if a function is technically continuous it will appear discontinuous when viewed at this level (in some sense no computational approximation is ever continuous, but we can observe the size of changes given a minimum resolution change). we can define this as follows (as in \cite{ilyas2019adversarial});

\begin{definition}{$\rho$-useful}
\end{definition}
For a given distribution $\mathcal{D}$, we call a feature $f$ $\rho$-useful ($\rho > 0$) if it is correlated with
the true label in expectation, that is if

\begin{align}
    \mathbb{E}_{(x, y) \sim \mathcal{D}}[y\cdot f(x)] \geq \rho
\end{align}
We then define $\rho_\mathcal{D}(f)$ as the largest $\rho$ for which feature f is $\rho$-useful under distribution $\mathcal{D}$.

\begin{definition}{$\gamma$-robustly useful}
\end{definition}

Suppose we have a $\rho$-useful feature $f (\rho 
 _D(f) > 0)$. We refer to $f$ as a robust feature if, under adversarial perturbation, f remains $\gamma$-useful. Formally, if we have that
\begin{align}
    \textrm{inf}_{\delta \in \Delta}\mathbb{E}_{(x, y) \sim \mathcal{D}}[y\cdot f(x + \delta)] \geq \gamma
\end{align}

\end{document}